\documentclass[final]{cvpr}

\usepackage[flushleft]{threeparttable}
\usepackage{multirow}
\usepackage{times}
\usepackage{epsfig}
\usepackage{graphicx}
\usepackage{amsmath}
\usepackage{amssymb}

\usepackage[pagebackref=true,breaklinks=true,colorlinks,bookmarks=false]{hyperref}

\def\cvprPaperID{52} 

\def\cvprPaperID{52} 
\def\confYear{WiCV 2021}

\begin{document}

\title{Robust Neonatal Face Detection in Real-world Clinical Settings}  

\author{Jacqueline Hausmann
\and
Md Sirajus Salekin
\and
Ghada Zamzmi
\and
Dmitry Goldgof
\and
Yu Sun\\
University of South Florida, Tampa, FL, United States\\

}
\maketitle


\begin{abstract}
   
    Current face detection algorithms are extremely generalized and can obtain decent accuracy when detecting the adult faces. These approaches are insufficient when handling outlier cases, for example when trying to detect the face of a neonate infant whose face composition and expressions are relatively different than that of the adult. It is furthermore difficult when applied to detect faces in a complicated setting such as the Neonate Intensive Care Unit. By training a state-of-the-art face detection model, You-Only-Look-Once, on a proprietary dataset containing labelled neonate faces in a clinical setting, this work achieves near real time neonate face detection. Our preliminary findings show an accuracy of 68.7\%, compared to the off the shelf solution which detected neonate faces with an accuracy of 7.37\%. Although further experiments are needed to validate our model, our results are promising and prove the feasibility of detecting neonatal faces in challenging real-world settings. The robust and real-time detection of neonatal faces would benefit wide range of automated systems (e.g., pain recognition and surveillance) who currently suffer from the time and effort due to the necessity of manual annotations. To benefit the research community, we make our trained weights publicly available at github\footnote{ \url{https://github.com/ja05haus/trained_neonate_face}}.
\end{abstract}

\section{Motivation and Challenges}

After the early 2000's, with the emergence of large and openly sourced image classification databases such as ImageNet \cite{imagenet_cvpr09} and Microsoft's COCO \cite{coco}, numerous accurate and real-time object detection algorithms were created and released. These algorithms are implemented using convolutional neural networks (CNN) and a variety of different factors, balancing speed and accuracy. Extremely generalized however, these solutions currently still face challenges resulting in low detection rates when taken directly off the shelf and employed in practical applications. 

This work focuses on face detection, an application widely used in several computer vision applications. More specifically, this work focuses on the detection of the neonatal face within the Neonatal Intensive Care Unit (NICU) setting.
General off the shelf solutions (e.g., R-CNN \cite{rcnn}) are currently knowledgeable about classes such as "person" as opposed to "face". Classes known to the model are determined by defined labels during training. In part due to the lack of accessible data, even deep learning models that understand the "face" class and can highly detect the human adult face within a frame are unable to detect the neonate face. This work shows that YOLO (You-only-look-once), which was introduced by Joseph Redmond \cite{yolov1}, can be adjusted and fine-tuned to detect neonatal face with high accuracy and promising near real time speeds (in some cases as fast as 8 fps, with an average detection rate of 4 fps).

The NICU is a challenging real-world environment for a variety of reasons. For example, the background scenes are often times highly complex containing overlapping wires and machines of similar colors as well as brightly colored patterned fabrics introducing noise.
Similarly, the neonate's face is often obscured by wires, oxygen masks, tape, or clothing. 

These factors make it difficult to identify face regions apart from background, causing many off the shelf face detection algorithms to fail. Additionally, limited space around the neonate while in the incubator and frequent changes in camera location lead to extreme pose variations. This is an additional challenge when attempting to detect the neonate face in the NICU.

While several state-of-the-art face detection approaches have taken into consideration general computer vision difficulties described above, there is another added challenge when attempting to detect the face of a neonate compared to the adult human face. Face-Channel \cite{faceChannel} presents a light-weight deep learning approach of adult face detection. \cite{9044721} is one of the first approaches specifically focusing of neonate face detection but looks to solve problem of pose variations.  O'Neil et al. \cite{13883176320191001} demonstrates the large inter-class variation between face expressions as a reaction to pain when considering infants at 2, 4, 6, and 12 months of age. As this work focuses exclusively on the neonate, and therefore no more than 4 weeks post birth, there is an even wider interval in terms of facial symmetry compared with an adult face. Facial features of infants are different from adults in a variety of ways including the lack of eyebrows, not fully developed facial bones leading to softer features, and compressed face dimensions truncating or squishing the eyes, nose, and mouth. All of these factors lead to difficulties in facial feature extraction key for an accurate face detection, demonstrated by the need for separate neonatal facial expression coding in \cite{1988-16499-00119870301}.  If there were possibly more publicly accessible datasets containing labelled neonate faces, some of these challenges could be overcome. Data of this type is hard to collect partly due to privacy concerns and the concept of unwilling/uncooperative participants.

\section{Neonatal Face Dataset and Preprocessing}

This section describes the dataset used in experiments throughout this work. Additionally, any preprocessing done upon the data in order to have viable testing and training sets is mentioned in section 2.2. 

\subsection{Neonatal Face Data}
The University of South Florida Multimodal Neonatal Pain Assessment Dataset (USF-MNPAD-I) \cite{dib_data} is the main dataset used throughout this work. USF-MNPAD-I was collected as the first iteration of a multi-modal neonatal dataset for a pain study presented by Zamzmi \cite{ghadadis}. Her work has evolved into an larger project with a focus on deep learning approaches to pain prediction in the neonate. This is a collaborative study between USF's Computer Science and Engineering Department, USF Health, and Tampa General Hospital. For more information, the project website is located at: \footnote{\url{https://rpal.cse.usf.edu/project_neonatal_pain/}}.
 
USF-MNPAD-I is composed of a total of 58 neonates (52.73\% female) with a gestational age range of 27 to 41 weeks and a mean birth weight of 2823.24 grams. The biggest in-balance comes from racial subsets with only 20\% of the subjects categorized as African American and 7.27\% as Asian, the rest Caucasian. The data was collected during the neonates' hospitalization at the NICU in Tampa General Hospital. Consent forms are obtained from the parents before data collection. 
 
 As a multi-modal dataset, USF-MNPAD-I has video, audio, cortical activity, and vital signs. Video data were recorded using of a GOPro RGB camera, with 1080p resolution and 30 FPS frame rate.  Concurrently, ambient room audio is collected through the GoPro Hero camera at 48KHz. As this work is explicitly dealing with face detection, only the video modality is considered. 
 
 The USF-MNPAD-I dataset was collected during two different types of pain: procedural and postoperative. The procedural pain is a short observable pain directly resulting from an external painful stimuli such as heel stick and neonatal immunization. Postoperative pain is longer, while a less intense type of pain, seen as a result to injury of internal tissue generally after a medically necessary surgery. Often times it is necessary to administer opioid medications to mitigate this type of pain which can lead to long term developmental effects \cite{S235255681930075X20190801, vdc.100075151526.0x00000120180101}. During the data collection of both types of pain, ground truth pain scores were provided by an expert nurse using clinical pain scales.

\begin{figure}[h]
\centering
\includegraphics[width=50pt,height=80pt]{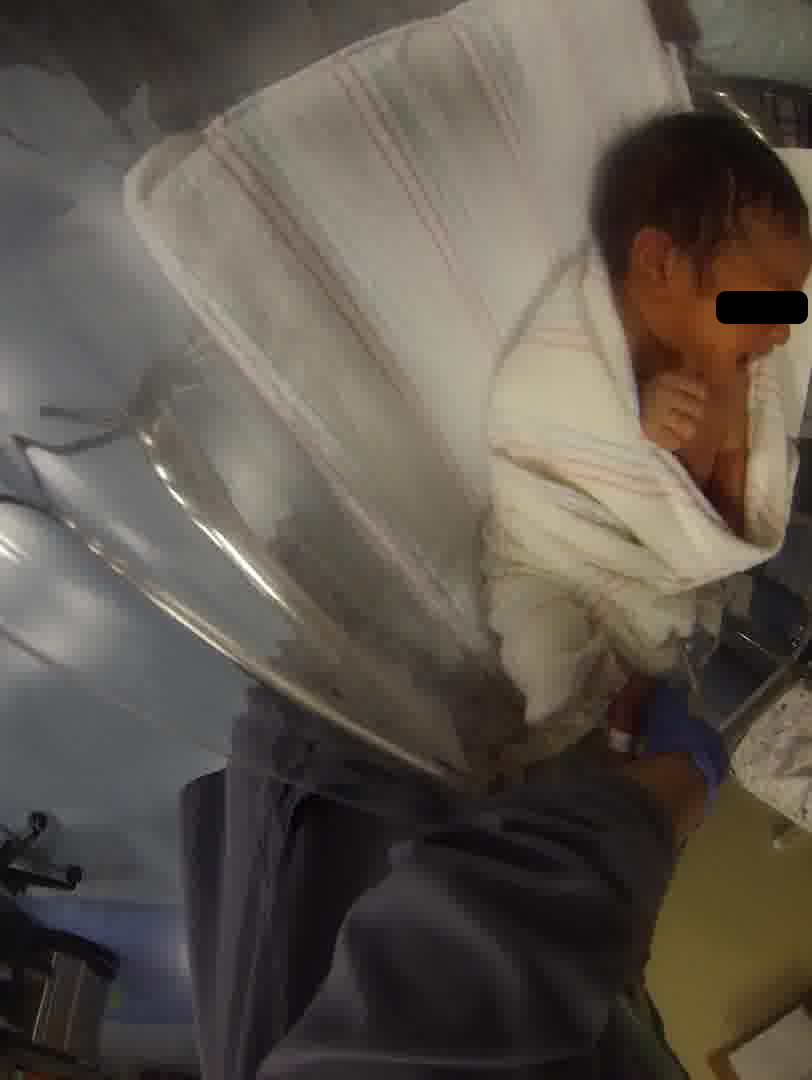}
\includegraphics[width=50pt,height=80pt]{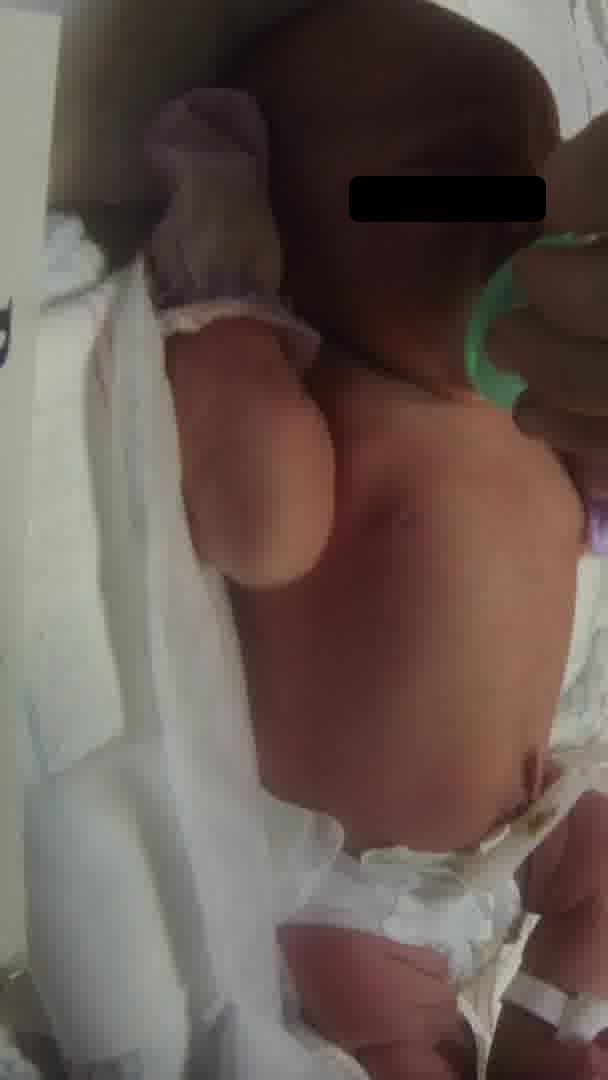}
\includegraphics[width=50pt,height=80pt]{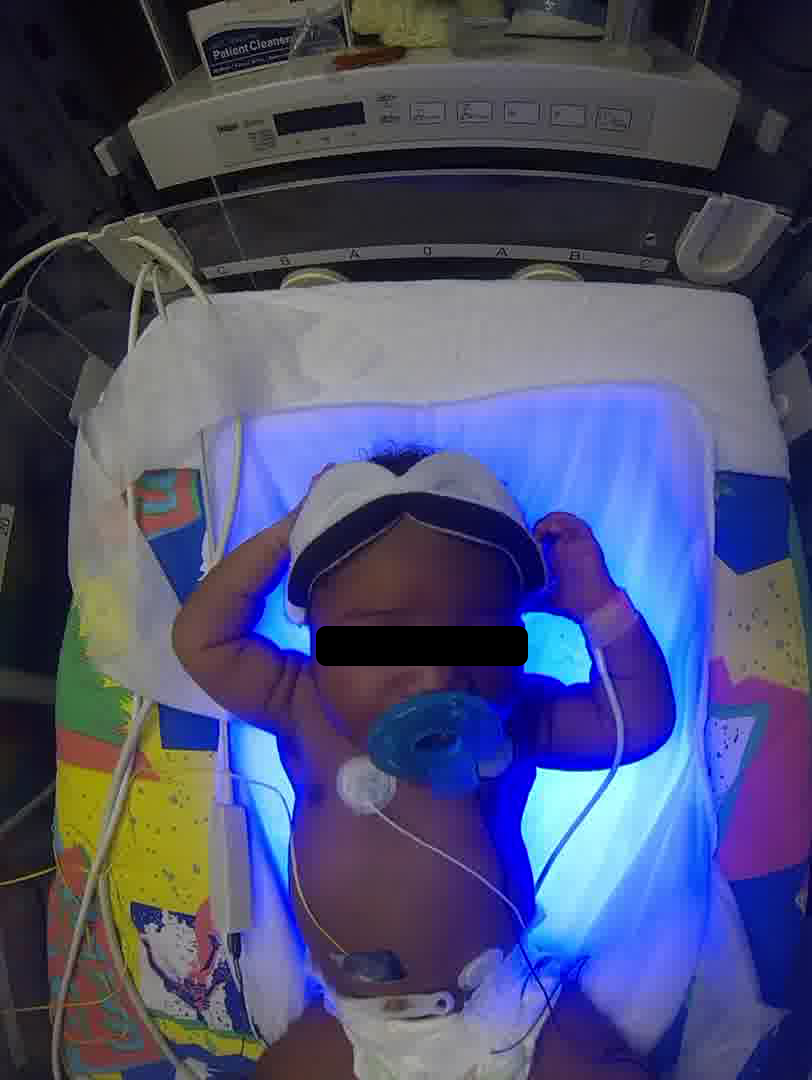}
\includegraphics[width=50pt,height=80pt]{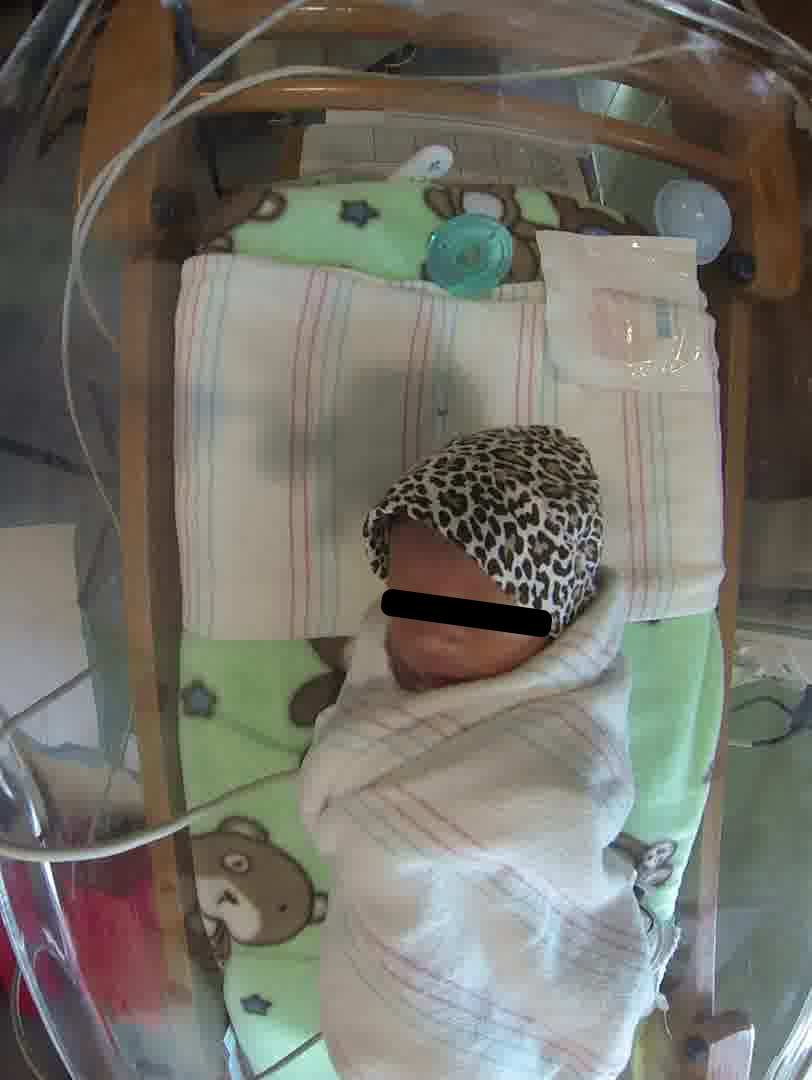}
\includegraphics[width=50pt,height=80pt]{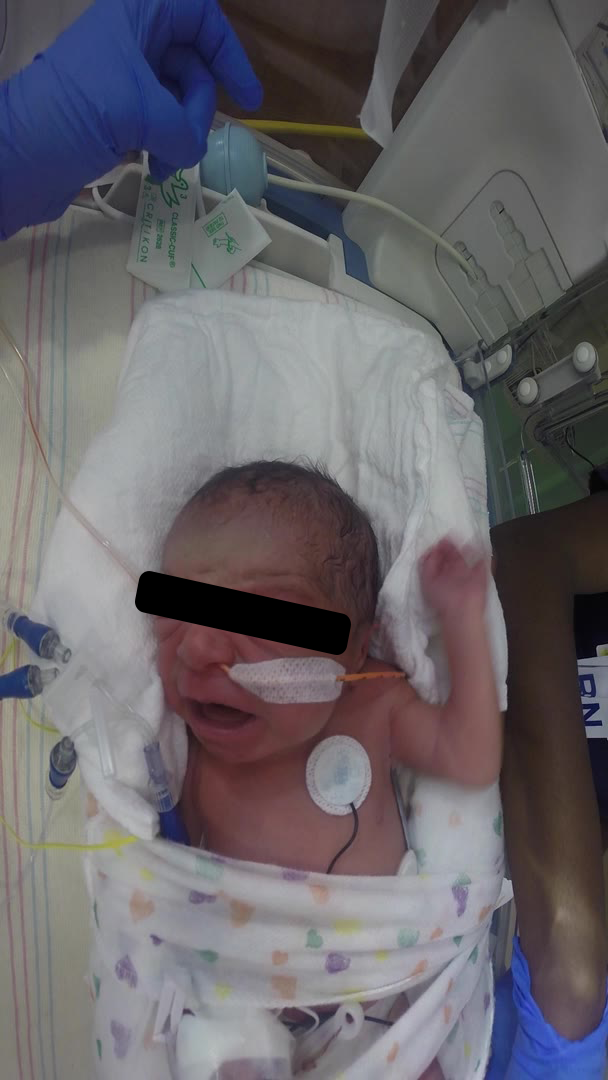} 
\includegraphics[width=50pt,height=80pt]{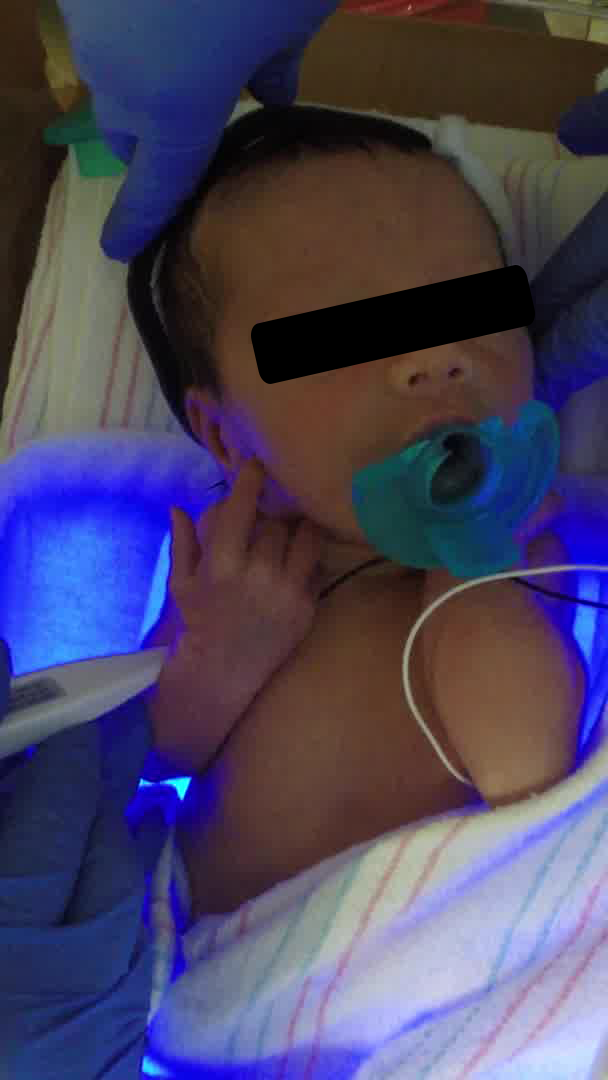}
\includegraphics[width=50pt,height=80pt]{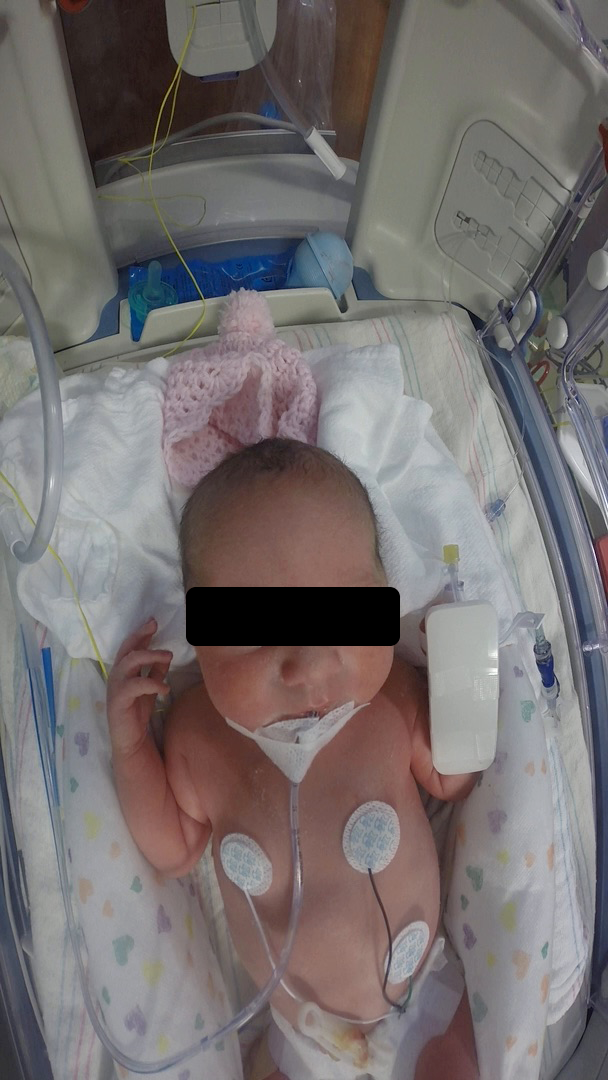} 
\includegraphics[width=50pt,height=80pt]{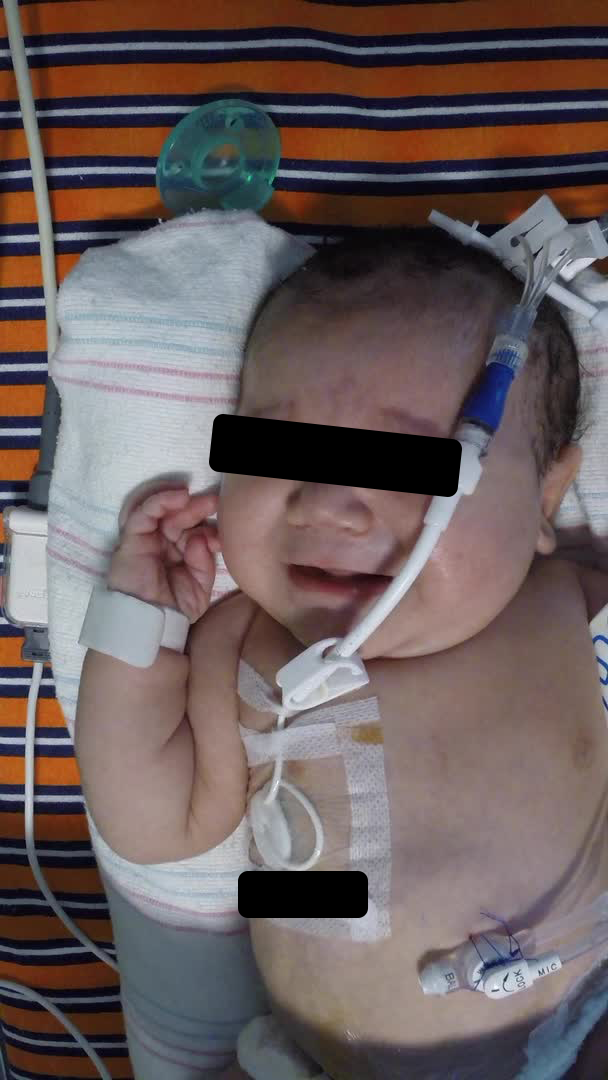}
\caption{Examples of USF-MNPAD-I images within NICU setting}
\end{figure}

Figure 1 presents a variety of sample images  demonstrating the challenging NICU environment, due to factors such as tape and wires obscuring the neonates face, low lighting conditions, complex background noise, etc. 
 
\subsection{Preprocessing}

In addition to the raw video files described previously, the labelled pain scores from a trained nurse was key for determining the source for images used throughout training and testing. While this work is not concerned with the overall pain classification of a frame, subsequent models used for multi-modal pain classification \cite{salekin2020investigation, edsarx.2012.0217520200101} will require the use of correctly detected neonatal faces and therefore will need a balance of frames labelled as no-pain/pain. This may not be the optimal final solution for training and testing image set composition, but at this time was maintained for consistency and in order to help eliminate adding human bias. 

\subsection{Training}

We used the procedural subset of USF-MNPAD-I, which has 36 subjects, for training the neonatal face detector. As these subjects were filmed during short procedures, video segments ranged from 1 min to 6 minutes in length. The Key frames of these video segments were extracted to obtain a total of 8,826 training images, with a per subject mean of 245 images.  Its worth noting, while face detection is not concerned with pain/no-pain classification labels, during the pain segments more variety in facial expressions is seen strengthening the variance in our training data.   

\subsection{Testing}

To measure the accuracy if a trained face detector, unseen testing set should be used. Hence, we evaluated our detector, which was trained on the procedural set of USF-MNPAD-I, on the postoperative set of USF-MNPAD-I.

As long hours were filmed per subject while recording the postoperative pain, the recorded raw video was sub-sampled according to marked pain/no-pain points in the ground truth timeline. With a total of 8 subjects, 512 different key frames were extracted per subject. This makes a total of 4,096 images in the test set. 

\subsection{Wider-Face}

As a baseline, we used YOLOv3 neonatal face detector \cite{8914537,salekin2020investigation}, which was fine-tuned on WIDER FACE \cite{widerface}. As such, it is necessary to consider what data comprises the WIDER FACE dataset, and how it differs from our testing and training sets of USF-MNPAD-I.

Wider Face is a large scale dataset with 32,203 total images containing 393, 703 faces which have been identified and labelled \cite{widerface}. It has a variety of depths, face postures, resolutions, backgrounds, and subjects making it an ideal dataset for training a generalized adult face detector. As discussed earlier, face detection of the neonate presents these challenges when recorded in the NICU. However, neonates have relatively different facial structures \cite{1988-16499-00119870301, 13883176320191001} as compared to the adults. Therefore, using a detector trained on Wider-Face, which only contains images of adults or older children, is insufficient, and might fail when applied to detect neonate face in practice.

\subsection{MS COCO}

For all the off the self comparisons, YOLOv5 \cite{v5code} is  pre-trained on Microsoft's Common Objects in Context \cite{coco}. This dataset has over 200k labelled images, with over 330k total images and 80 object classes. It is a very generalized dataset, with no "face" class only "person". It provides a reasonable starting point, but overall accuracy using this dataset is not sufficient for this application.

\section{Methodology}

This section provides an in-depth description of YOLO, a CNN based object detection model known for its speed and accuracy in bounding box detection. In this work, we used the most recent version (V5) as the base model for face detection. This version is publicly available for research use from Ultralytics \cite{v5code}.

\subsection{Bounding Box Detection}
Given an image, YOLO splits the image into an \textit{S} x \textit{S} grid. Each of the individual boxes of the grid will be responsible for detecting the object in the image whose center falls within a respective box. 
Each individual cell in the \textit{S} x \textit{S} grid will compute \textit{B} bounding boxes, as well as a confidence score for \textit{B} bounding boxes. Confidence of a bounding box, (given object \textit{O}, predicted bounding box \textit{A}, ground truth bounding box \textit{B}), is defined as the 
    \begin{gather}
        P(O) * \frac {A \cap B}{ A \cup B}
    \end{gather} 
    
That is the probability of a given object multiplies by the intersection over union between predicted and ground truth bounding boxes. The cells within the S x S grid will have only have a single set of corresponding class probabilities, given \textit{C} classes \cite{yolov1}. This additional variable multiplied by the per box bounding confidence, results in a final class dependent confidence score.
    
The supplementary values contained within each \textit{B} bounding boxes include \textit{x, y, w,} and \textit{h}. \textit{x} and \textit{y} identify the center point of each bounding box, while \textit{w} and \textit{h} are weight and height, respectively, of each specific bounding box normalized by full image width and height to the [0-1] range. 

\subsection{You Only Look Once Architecture} 

YOLO can be described as a classical convolutional neural network. Initially with an overall total of 24 convolutional layers in YOLOv1, first a $7$x$7$ convolutional layer is applied to the input image. Following the first convolutional layer, are repeated $3$x$3$ convolutional layers and then a $1$x$1$ reduction layer used for feature space reduction. The resulting tensor from previous convolutional layers is then passed through two fully connected layers with a final parameterized space of $S$x$S$x$(B *5 + C)$. The 5 is due to the constants corresponding to the 5 individual variables that make up a bounding box prediction described in the previous section.
    
By YOLOv3 detection speeds had been reduced to 22ms on a Titan X, incorporating a larger architecture while producing higher mean average precision. This new network includes 53 convolutional layers (compared to YOLOv1's 24), garnering it the name Darknet-53 \cite{yolov3}. Still included are the repeated $3$x$3$ convolutional layers, then $1$x$1$ convolutional layers. Additionally, 3x3/2 convolution layers are included as a way to implement shortcut connections. Darknet-53 remains the backbone for all current versions of YOLO, although by YOLOv4 an added neck and head component to the architecture takes the forefront \cite{yolov4}. Darknet-53 is pre-trained on ImageNet \cite{imagenet_cvpr09}, with a head component responsible for actual class determination and bounding box coordinates. YOLO is considered a leading one-stage object detector. 

\subsection{Advancements in Speed \& Accuracy}

After YOLOv3, Darknet-53 retained as the base architecture for all future versions of YOLO. Cross Mini-Batch Normilization (CmbN) is an optimized batch normalization approach in which mean and variance values are updated only after 4 mini batches. Bochkovskiy et al\ cite{yolov4} show that implementing CmbN will decrease inference processing time by 12\% fps processing time. 

Considering the composition of an image, it is often useful to focus on spatial relationships between output features from convolutional layers for increased accuracy. Attention modules such as Squeeze-and-Excited (SE) \cite{squeeze} and Spatial Attention Module (SAM) \cite{sam} present the fore front approaches to this concept. YOLOv5 uses a modified SAM model \cite{yolov4} for spatial emphasised feature extraction. When used with a ResNet50 base model, SE is shown to have increase in 1\% top ten accuracy on ImageNet \cite{imagenet_cvpr09} with the cost of 10\% per inference time \cite{yolov4}. In contrast, SAM saw an increase in 0.5\% accuracy, while only costing 0.1\% additional per inference time. Utilizing a modified SAM which applies concatenation between average pooling and max pooling feature maps, allows YOLOv5 to increase accuracy with little to no additional computational cost.  For any system to work in a real world setting, face detection speeds must approach real time. With an average of 0.27 seconds (total of 1108.681 seconds for 4096 images) on the test set, this work shows that this is near possible. This detection speed may be increased in the future when working with temporal video data, as not every frame may need to be analyzed.  

Maintaining near real time detection speeds through the use of tools described above, YOLOv5 utilizes multiple different deep learning optimizations for increase detection accuracy. While it has been shown that regularization techniques such as dropout and weight decay will help reduce overall parameters while deep neural networks' learn, due to the spacial relationship inherent in image, convolution neural networks don't normally see as much of a benefit from implementing those techniques. Introduced in 2018 by Ghiasi et al \cite{dropbox}, DropBox looks to overcome the inept performance when applying dropout to convolutional layers. Dropbox injects noise to larger blocks of area instead of individual activation units. Dropbox will be implemented beginning with YOLOv4 \cite{yolov4} due to its consistent increased performance when used in convolutional neural network layers.

Another huge increase in accuracy came through the use of Mosaic augmentation. It has been previously shown that data augmentation allows for most robust and accuracy image detection systems, Mosaic augmentation capitalizing upon this theory. Based off Cutmix \cite{cutmix}, Mosaic augmentation is run time augmentation which combines multiple training images into a single input to the model. By combining multiple images into one with Mosaic augmentation, the network is provided a larger range of background context, scales, rotations, and data transformations resulting in a robust network. Ultimately this translates into a higher accuracy, especially in difficult clinical settings.

\section{Results of Trained YOLOv5}

This section reinforces that using YOLOv5 trained on the procedural subset of USF-MNPAD-I would provide an accurate and efficient neonatal face detection solution. 

\subsection{Training Hyper parameters and Environment}

All of the training and testing for this paper was done on a GPU cluster maintained by the University of South Florida's College of Engineering 
for student research purposes. This server uses CentOS 7.4 as its operating system and is composed of GeForce GTX 1080 Ti GPU nodes. Training of YOLOv5(s, the smallest version) on the GPU cluster took an average of 22.5 hours for 1000 epochs.  

The specific hyper parameters used for training are as follows: image input size $416 \times 416$, batch size of 16, 1000 epochs in length, learning rate 0.01, momentum 0.937, weight decay 0.0005, with a warm-up time of 3 epochs and a warm-up momentum of 0.8. In addition to Mosaic data augmentation, random translations of 0.1 and random scaling at 0.5 are implemented. 

\subsection{Results}

Our results show that the trained YOLOv5 is able to detect the neonatal face within the difficult clinical NICU setting at an improvement of more than 61\% when compared to the off the self solution. 

Firstly, a ten-fold cross validation approach was done during training, in order to prove statistically significant improvements were possible with this approach. With a total of 36 training subjects, maintaining subject boundaries this means 3 subjects reserved for testing each fold, with 4 subjects left out in the final fold so that each subject is used for testing at least once during cross validation folds. We saw 99\% accuracy for cross-validation training results with YOLOv5 trained on the procedural subset of USF-MNPAD-I. This is compared to our baseline of 61.7\% when using YOLOv3 trained on the adult faces in WIDER-FACE \cite{widerface}, and 31.7\% when using YOLOv5 with no additional training. 

For all of our results, the bounding box must have a confidence of high that 40\% to be considered. This was an empirically dervied threshold observed to be fair during training and testing. 

Furthermore, the IOU of ground truth bounding box to predicted bounding box must have a higher value than 0.5, otherwise the detection was deemed inaccurate. The S\o rensen-Dice Coefficent (SDC) was used as the metric to quantify the similarities between the predicted and ground truth samples. Mathematically described to be applied to the bounding box problem, the SDC is represented by four metrics, True Positive ($TP$), True Negative ($TN$), False Positive ($FP$), and False Negative ($FN$). 

\begin{gather}
    SDC = \frac{2TP}{2TP + FP + FN}
\end{gather}

After a 10 fold cross validation approach verified the increase in accuracy of this approach, the trained model is evaluated on a testing set. Similarly, an SDC of 0.3 was allowed, but maintained a threshold confidence score of 0.4. These results of face detection on the testing set are reported in Table 1.

\begin{table}[h]
    \begin{center}
        \caption{Performance of Face Detection on Test Set (postoperative set). The Totals indicates the total number of subjects in the testing set (8). YOLOv5-T indicates the model is trained on the procedural set.}
        \begin{tabular}{|c|c|c|} 
            \hline
            Subject & YOLOv5-NT & YOLOv5-T\\
            \hline
            \hline
            1 & 142 & 22 \\
            8 & 0 & 488 \\
            9 & 0 & 472 \\
            16 & 3 & 254 \\
            34 & 0 & 325 \\ 
            38 & 21 & 342 \\
            43 & 136 & 512 \\
            51 & 0 & 399 \\
            \hline
            \multirow{2}{*}{Totals} & 302 (images) & \textbf{2814} (images) \\
                & 7.37\% & \textbf{68.7\%} \\
            \hline
        \end{tabular}
    \end{center}
\end{table}

\begin{figure}[h]
\centering
\includegraphics[width=40pt,height=65pt]{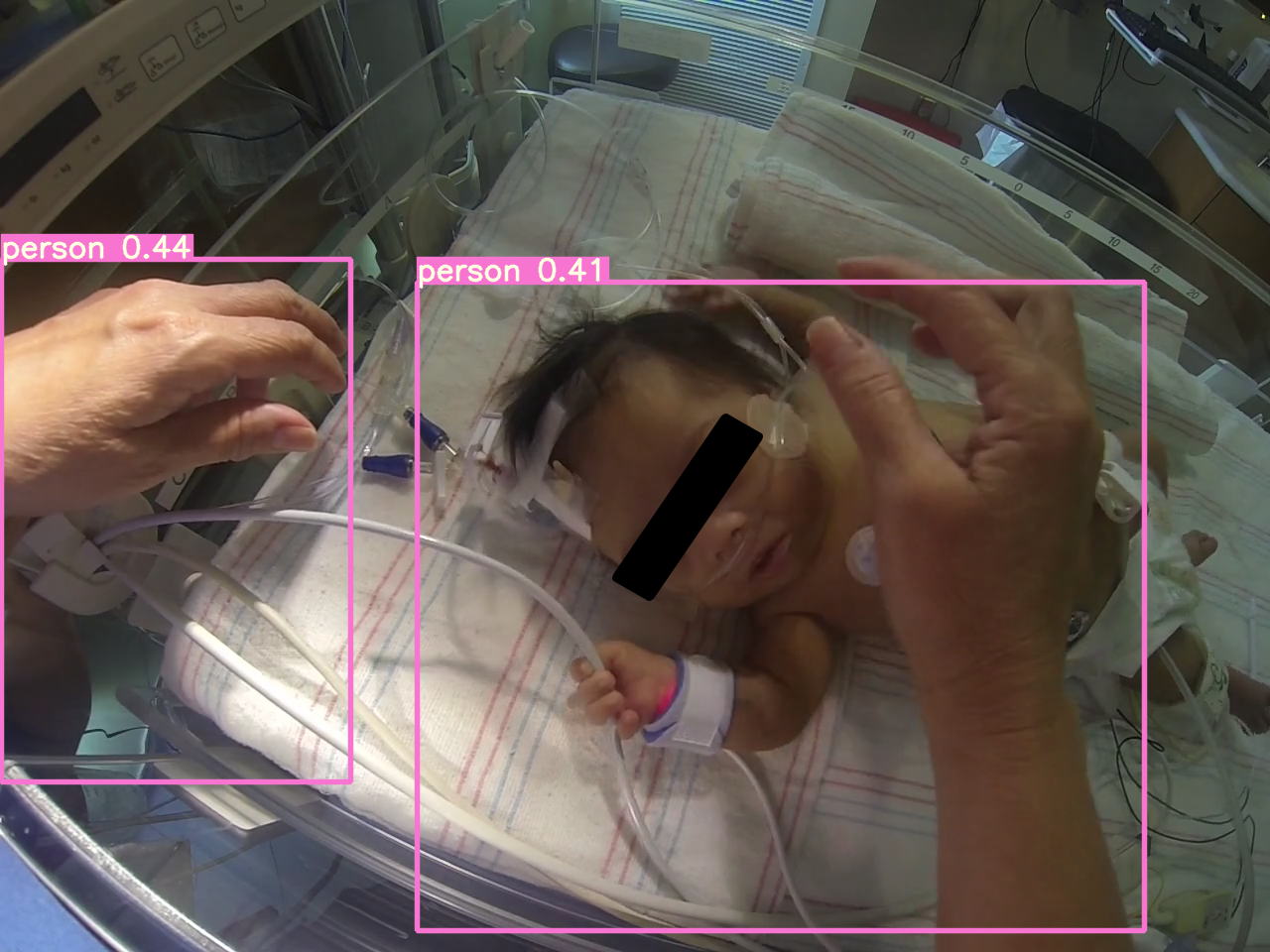}\qquad
\includegraphics[width=40pt,height=65pt]{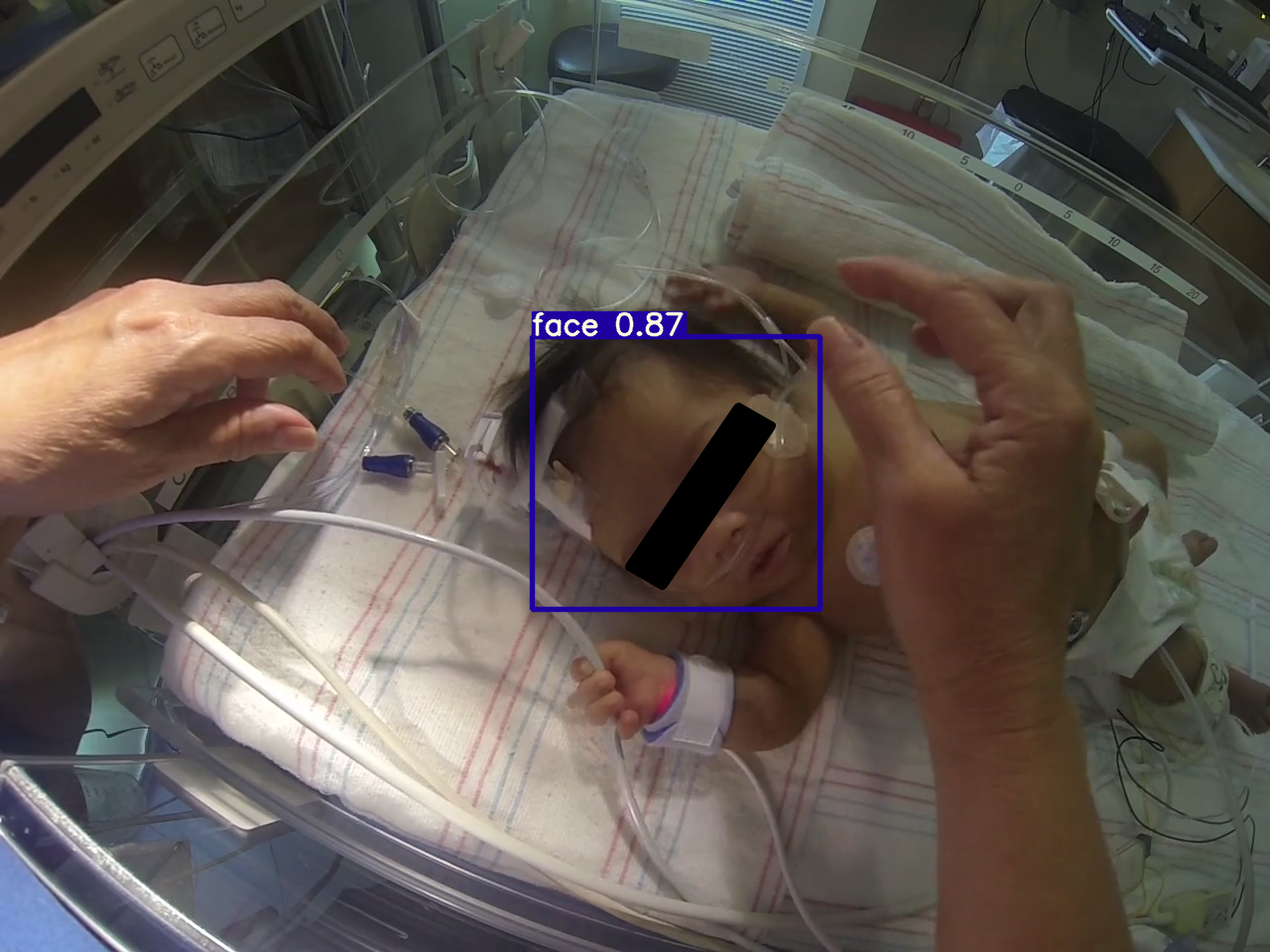} \qquad
\includegraphics[width=40pt,height=65pt]{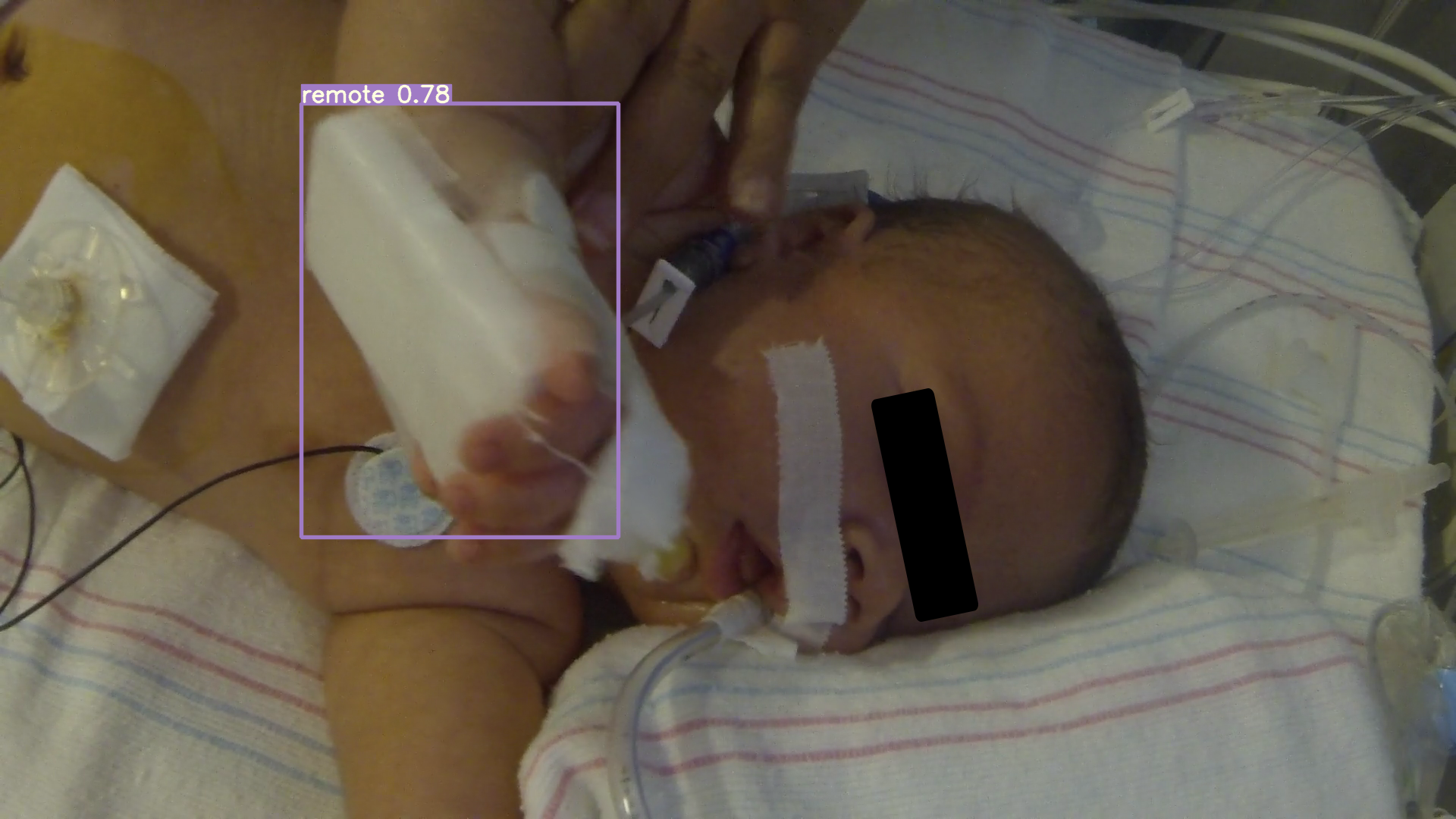} \qquad
\includegraphics[width=40pt,height=65pt]{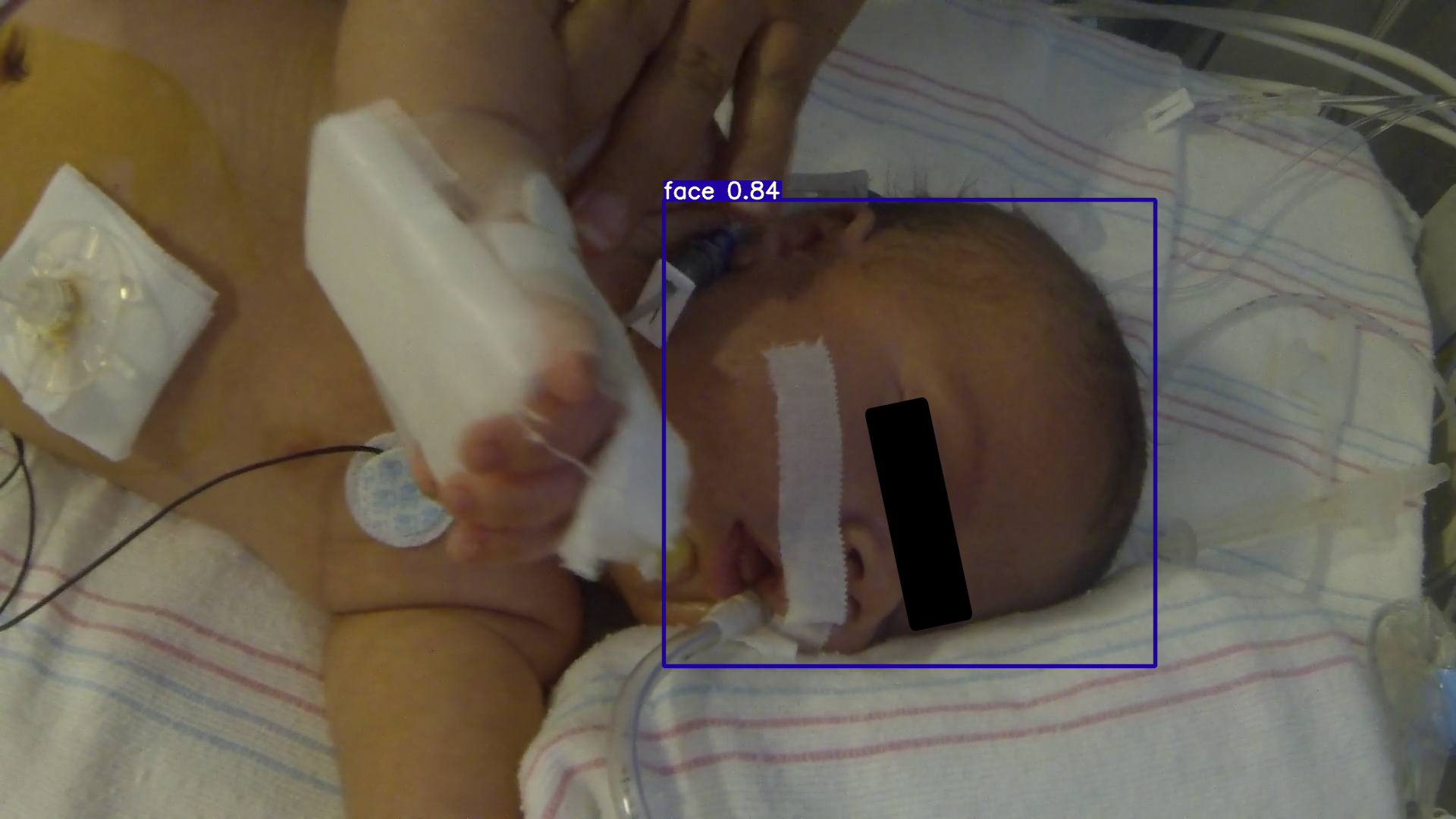} \qquad
\includegraphics[width=40pt,height=65pt]{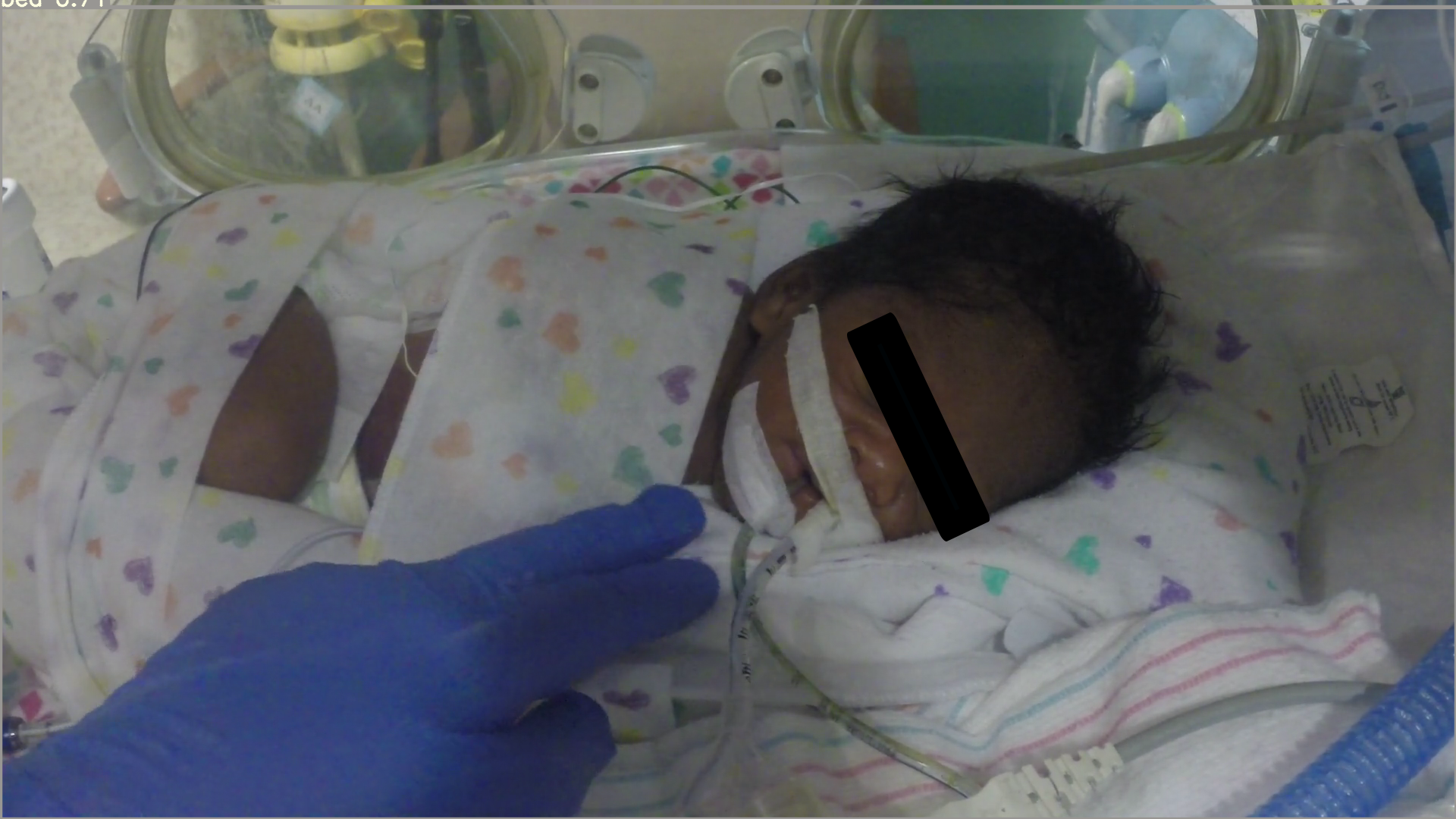} \qquad
\includegraphics[width=40pt,height=65pt]{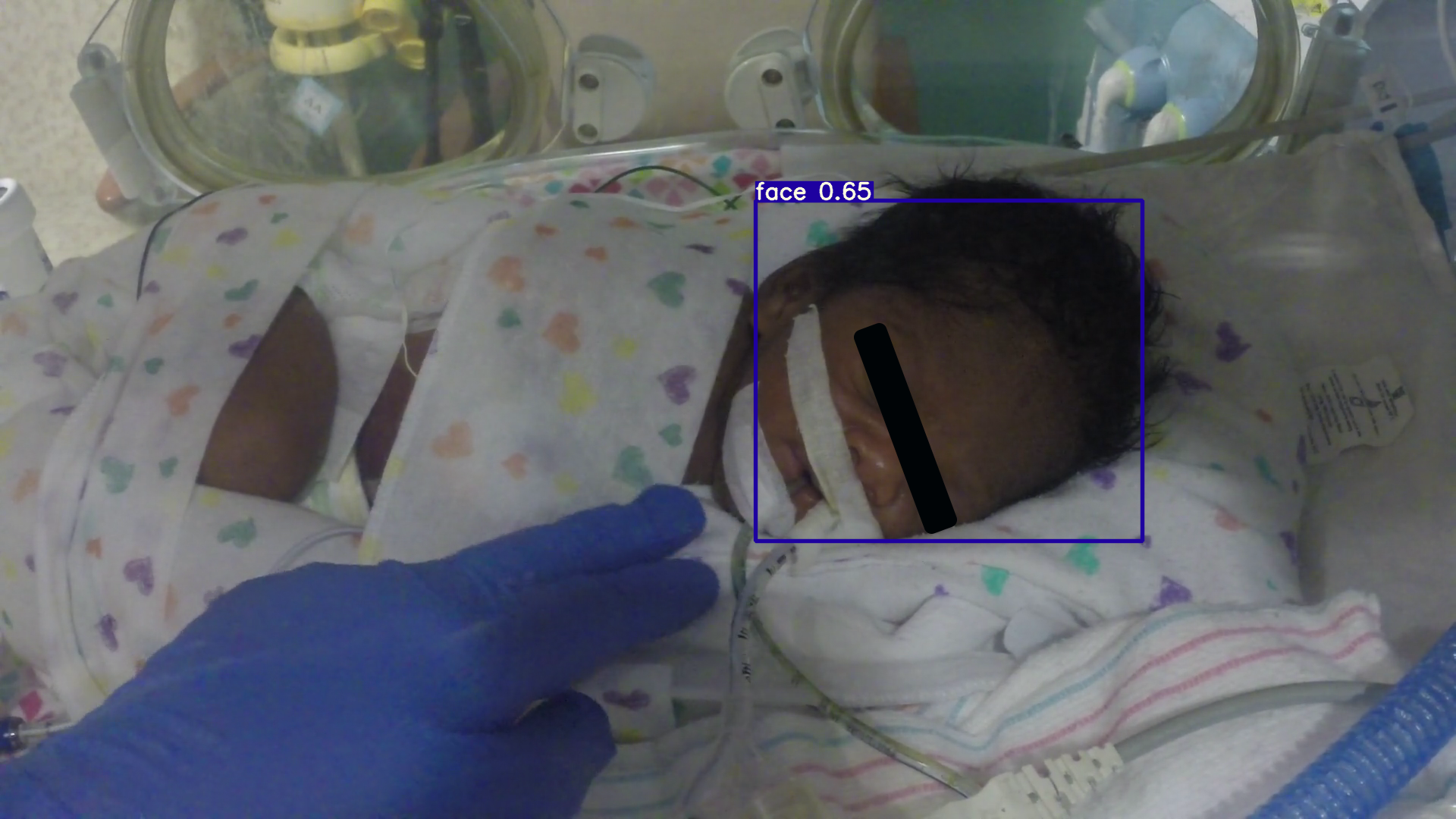} \qquad
\includegraphics[width=40pt,height=65pt]{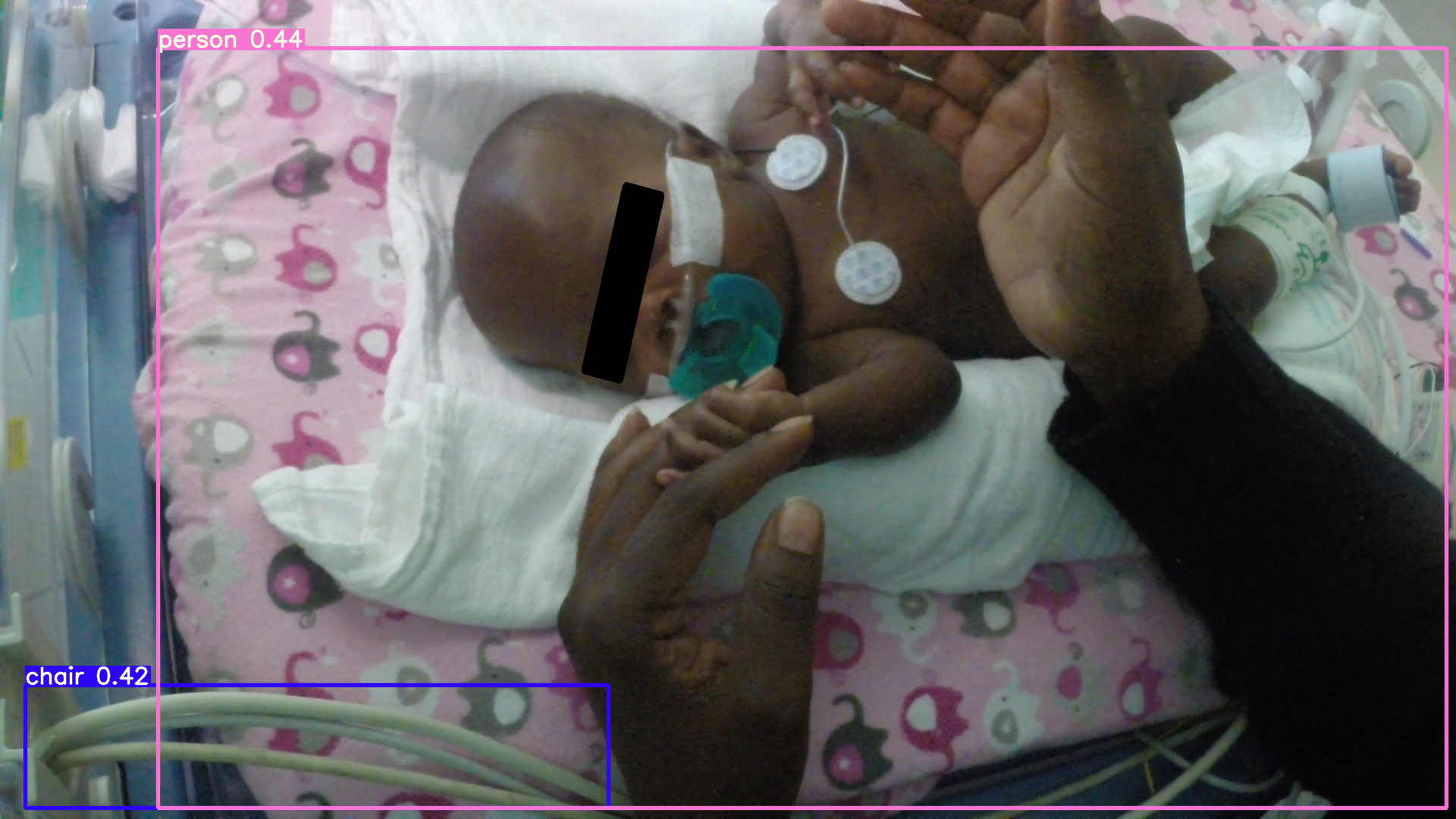} \qquad
\includegraphics[width=40pt,height=65pt]{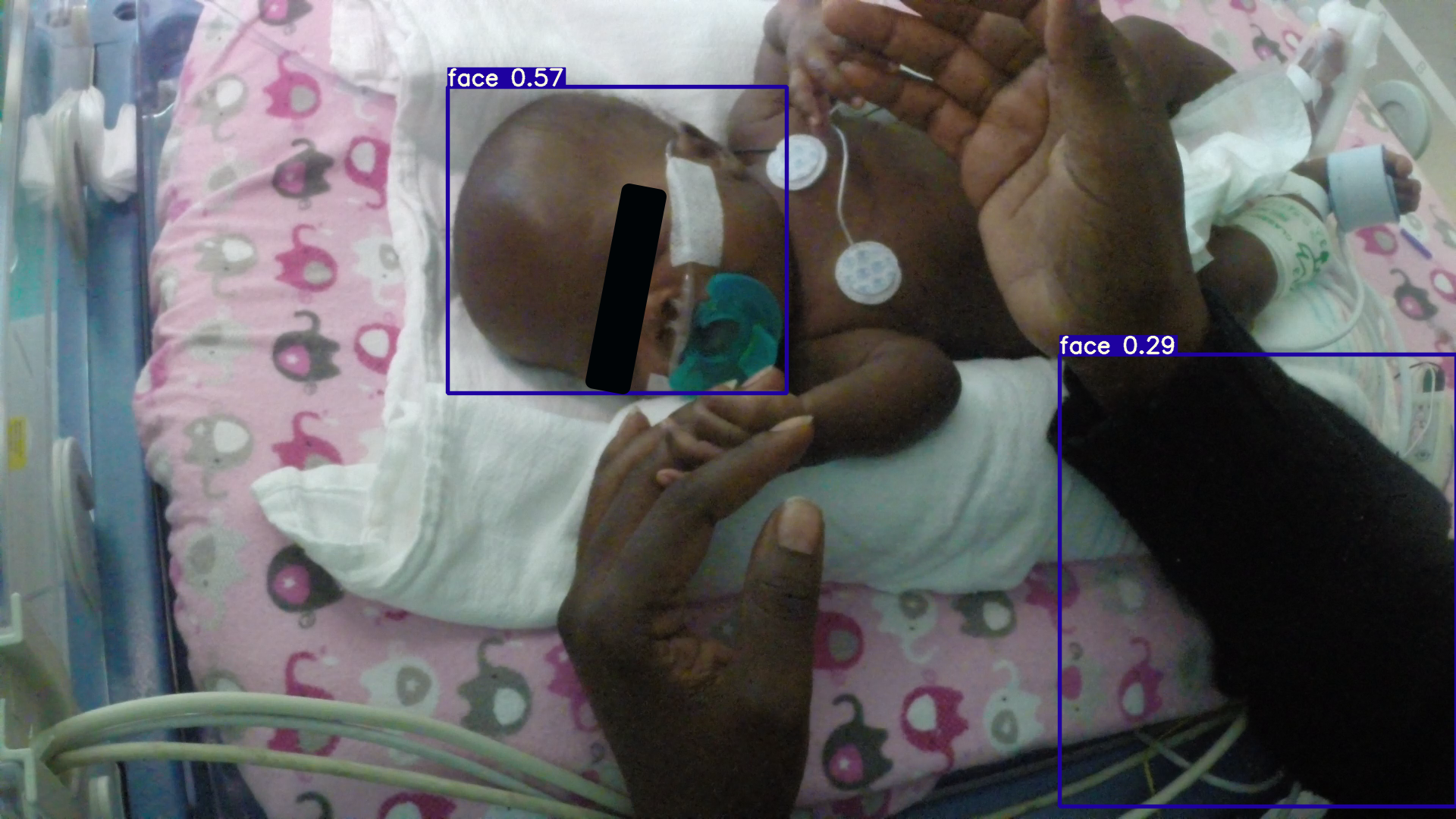} \qquad
\caption{Comparison of NT and Trained YOLO output}
\end{figure}
 
Figure 2 shows examples of the differences between outputs. On the left is the output from YOLOv5 with no additional training, with the results of face detection using trained YOLOv5 of the same image on the right. In many instances, YOLOv5 not trained will identify non relative classes within the scene such as "chair" or "remote". Even in the cases where the neonate is detected as "person"
the face is not identified, instead the bounding box encompassing the entirety of the neonate and often the hands of a caretaker. In each of these cases, the trained YOLOv5 is able to correctly identify the neonate face within a tightly localized bounding box with a high class confidence. Done at an average of 4 fps, this indicates in a real-world clinical setting accurate neonatal face detection is possible with our proposed model.

\section{Conclusion}
By training YOLOv5 on the procedural subset of USF-MNPAD-I, we are able to get near 70\% overall accuracy in neonatal face detection in a challenging NICU environment. In a very few instances, if the neonate's face composing a majority of the scene then a generalized "person" detector is able to detect the neonate within the scenes, but these are outlier cases and are insufficient for real-world clinical usage. 

In addition to pain detection in the clinical setting, this model has potential value in eliminating manual annotations for other applications. For example, \cite{7736912, faceRecDissertation, ICPRFace} are works experimenting with automating neonate facial recognition systems who suffer from the necessity of manual face detection. 

\subsection{Future Work}

As future work, we plan to measure how the proposed face detection model can help in pain detection, though there are a variety of applications which would benefit from the use of an accurate neonatal face detection system. In terms of implementing a better face detection model, many approaches such as ensemble learning could help improve the accuracy. Additionally, as data collection on USF-MNPAD-II is currently underway, adding post-operative subjects to the training set supplementary to the procedural pain subjects, will increase overall neonate face detection for future uses. 

\subsection{Impact in Clinical Setting}

The most important extension of this work is how it can actually be put into use in the clinical setting. While there has been an emphasis in recent years on increasing accuracy, these systems are tested in isolation. There is a gap between what nurses and doctors would be able to actually implement during daily monitoring and what computer scientist are able to detect using expensive computational heavy lab systems. Many bio-medical applications require systems that can be run without highly technically trained individuals. What's more, if we consider where is the need for these systems its not necessarily in state the art NICU's with many nurses and doctors on staff and large budgets. Creating solutions that could work in home or hospitals in impoverished neighborhoods, will allow healthcare systems whose capabilities are stretched to accurately automate minutia tasks. 

The difficulties seen in acquisition of data for USF-MNPAD-I highlight this. Using the most durable and compact on the market camera solutions, it has taken many iterations for prolonged video and audio acquisition. This doesn't account for human supervised preprocessing required to clean the data in order to be passed to any deep neural network system. This is a time consuming process which interferes with a systems ability to perform classification at real time. With a reliable and accurate neonate face detector, this process could be automated. It is a clear connection that these difficulties will manifest into larger roadblocks when considering the lengthy daily use of an all in one pain classification system. 

Many computer vision applications require face detection as the first step in data processing pipeline. This demonstrates the need for an accurate solution, but accuracy must not be sacrificed for efficiency. Currently, many approaches rely on lengthy human supervised manual annotation for face detection. Automating this process through the use of the proposed model would help begin to bridge the gap between theoretical solutions and solutions to real life clinical problems.  

\section*{Acknowledgement}
The work done in this paper is funded in part by National Institutes of Health Grant (NIH R21NR018756) and the University of South Florida Nexus Initiative (UNI) Grant.

{\small
\bibliographystyle{ieee_fullname}
\bibliography{references}

\begin{thebibliography}{10}\itemsep=-1pt

\bibitem{faceRecDissertation}
Mallika Agarwal.
\newblock Analysis and evaluation of algorithms for newborn face recognition,
  2017.

\bibitem{S235255681930075X20190801}
Fr{\'e}d{\'e}ric Aubrun, Karine Nouette-Gaulain, Dominique Fletcher, Anissa
  Belbachir, H{\'e}l{\`e}ne Beloeil, Michel Carles, Philippe Cuvillon,
  Christophe Dadure, Gilles Lebuffe, Emmanuel Marret, et~al.
\newblock Revision of expert panel's guidelines on postoperative pain
  management.
\newblock {\em Anaesthesia Critical Care \& Pain Medicine}, 38(4):405--411,
  2019.

\bibitem{9044721}
M. {Awais}, C. {Chen}, X. {Long}, B. {Yin}, A. {Nawaz}, S.~F. {Abbasi}, S.
  {Akbarzadeh}, L. {Tao}, C. {Lu}, L. {Wang}, R.~M. {Aarts}, and W. {Chen}.
\newblock Novel framework: Face feature selection algorithm for neonatal facial
  and related attributes recognition.
\newblock {\em IEEE Access}, 8:59100--59113, 2020.

\bibitem{faceChannel}
Pablo Barros, Nikhil Churamani, and Alessandra Sciutti.
\newblock The facechannel: A light-weight deep neural network for facial
  expression recognition.
\newblock April 2020.

\bibitem{7736912}
L. {Best-Rowden}, Y. {Hoole}, and A. {Jain}.
\newblock Automatic face recognition of newborns, infants, and toddlers: A
  longitudinal evaluation.
\newblock In {\em 2016 International Conference of the Biometrics Special
  Interest Group (BIOSIG)}, pages 1--8, 2016.

\bibitem{yolov4}
Alexy Bochkovskiy, Chien-Yao Wang, and Hong Yuan~Mark Liao.
\newblock Yolov4: Optimal speed and accuracy of object detection.
\newblock April 2020.

\bibitem{imagenet_cvpr09}
J. Deng, W. Dong, R. Socher, L.-J. Li, K. Li, and L. Fei-Fei.
\newblock {ImageNet: A Large-Scale Hierarchical Image Database}.
\newblock 2009.

\bibitem{dropbox}
Golnaz Ghiasi, Tsung-Yi Lin, and Quoc~V Le.
\newblock Dropblock: A regularization method for convolutional networks.
\newblock In S. Bengio, H. Wallach, H. Larochelle, K. Grauman, N. Cesa-Bianchi,
  and R. Garnett, editors, {\em Advances in Neural Information Processing
  Systems 31}, pages 10727--10737. Curran Associates, Inc., 2018.

\bibitem{rcnn}
Ross Girshick, Jeff Donahue, Trevor Darrell, and Jitendra Malik.
\newblock Rich feature hierarchies for accurate object detection and semantic
  segmentation.
\newblock pages 580--587, 2014.

\bibitem{1988-16499-00119870301}
Ruth~VE Grunau and Kenneth~D Craig.
\newblock Pain expression in neonates: facial action and cry.
\newblock {\em Pain}, 28(3):395--410, 1987.

\bibitem{squeeze}
Jie Hu, Li Shen, and Gang Sun.
\newblock Squeeze-and-excitation networks.
\newblock June 2018.

\bibitem{v5code}
Glenn Jocher.
\newblock Yolov5, Aug. 2020.

\bibitem{coco}
Tsung-Yi Lin, Michael Maire, Serge Belongie, James Hays, Pietro Perona, Deva
  Ramanan, Piotr Doll{\'a}r, and C~Lawrence Zitnick.
\newblock Microsoft coco: Common objects in context.
\newblock pages 740--755, 2014.

\bibitem{13883176320191001}
Monica~C. O'Neill, Sara Ahola~Kohut, Rebecca Pillai~Riddell, and Harriet Oster.
\newblock Age-related differences in the acute pain facial expression during
  infancy.
\newblock {\em European Journal of Pain}, 23(9):1596 -- 1607, 2019.

\bibitem{yolov1}
Joseph Redmon, Santosh Divvala, Ross Girshick, and Ali Farhadi.
\newblock You only look once: Unified, real-time object detection, 2016.

\bibitem{yolov3}
Joseph Redmon and Ali Farhadi.
\newblock April 2018.

\bibitem{8914537}
M.~S. {Salekin}, G. {Zamzmi}, D. {Goldgof}, R. {Kasturi}, T. {Ho}, and Y.
  {Sun}.
\newblock Multi-channel neural network for assessing neonatal pain from videos.
\newblock pages 1551--1556, 2019.

\bibitem{salekin2020investigation}
Md~Sirajus Salekin, Ghada Zamzmi, Dmitry Goldgof, Rangachar Kasturi, Thao Ho,
  and Yu Sun.
\newblock First investigation into the use of deep learning for continuous
  assessment of neonatal postoperative pain.
\newblock pages 415--419, 2020.

\bibitem{edsarx.2012.0217520200101}
Md~Sirajus Salekin, Ghada Zamzmi, Dmitry Goldgof, Rangachar Kasturi, Thao Ho,
  and Yu Sun.
\newblock Multimodal spatio-temporal deep learning approach for neonatal
  postoperative pain assessment.
\newblock {\em Computers in Biology and Medicine}, 129:104150, 2021.

\bibitem{dib_data}
Md~Sirajus Salekin, Ghada Zamzmi, Jacqueline Hausmann, Dmitry Goldgof,
  Rangachar Kasturi, Marcia Kneusel, Terri Ashmeade, Thao Ho, and Yu Sun.
\newblock Multimodal neonatal procedural and postoperative pain assessment
  dataset.
\newblock {\em Data in Brief}, 35:106796, 2021.

\bibitem{ICPRFace}
S. {Siddiqui}, M. {Vatsa}, and R. {Singh}.
\newblock Face recognition for newborns, toddlers, and pre-school children: A
  deep learning approach.
\newblock In {\em 2018 24th International Conference on Pattern Recognition
  (ICPR)}, pages 3156--3161, 2018.

\bibitem{sam}
Sanghyun Woo, Jongchan Park, Joon-Young Lee, and In~So Kweon.
\newblock Cbam: Convolutional block attention module.
\newblock September 2018.

\bibitem{widerface}
Shuo Yang, Ping Luo, Chen-Change Loy, and Xiaoou Tang.
\newblock Wider face: A face detection benchmark.
\newblock pages 5525--5533, 2016.

\bibitem{cutmix}
Sangdoo Yun, Dongyoon Han, Seong~Joon Oh, Sanghyuk Chun, Junsuk Choe, and
  Youngjoon Yoo.
\newblock Cutmix: Regularization strategy to train strong classifiers with
  localizable features.
\newblock pages 6023--6032, 2019.

\bibitem{ghadadis}
Ghada Zamzmi.
\newblock {\em Automatic Multimodal Assessment of Neonatal Pain}.
\newblock PhD thesis, University of South Florida, 2018.

\bibitem{vdc.100075151526.0x00000120180101}
Ghada Zamzmi, Rangachar Kasturi, Dmitry Goldgof, Ruicong Zhi, Terri Ashmeade,
  and Yu Sun.
\newblock A review of automated pain assessment in infants: features,
  classification tasks, and databases.
\newblock volume~11, pages 77--96. IEEE, 2017.

\end{thebibliography}
}

\end{document}